\documentclass[sigconf]{acmart}
\usepackage{times}
\usepackage{epsfig}
\usepackage{graphicx}
\usepackage{amsmath}
\usepackage{tikz}
\usepackage{pgfplots}
\usepackage{threeparttable}
\usepackage{multirow,multicol, makecell, booktabs}
\usepackage{threeparttable}
\RequirePackage{subfigure}
\usepackage{xcolor}
\AtBeginDocument{%
  \providecommand\BibTeX{{%
    \normalfont B\kern-0.5em{\scshape i\kern-0.25em b}\kern-0.8em\TeX}}}

\pgfplotsset{compat=1.17}
\begin{document}
\settopmatter{printacmref=false} 

\renewcommand\footnotetextcopyrightpermission[1]{} 
\pagestyle{plain} 
\title{Explore before Moving: A Feasible Path Estimation and Memory Recalling Framework for Embodied Navigation}

\author{Yang Wu}
\affiliation{%
  \institution{School of Computer Science and Engineering}
  \city{Sun Yat-sen University}
  \country{}
  }
\email{wuyang36@mail2.sysu.edu.cn}
\author{Shirui Feng}
\affiliation{%
  \institution{School of Computer Science and Engineering}
  \city{Sun Yat-sen University}
  \country{}
  }
\email{fengshr@mail2.sysu.edu.cn}
\author{Guanbin Li}
\affiliation{%
  \institution{School of Computer Science and Engineering}
  \city{Sun Yat-sen University}
  \country{}
  }
\email{liguanbin@mail.sysu.edu.cn}
\author{Liang Lin}
\affiliation{%
  \institution{School of Computer Science and Engineering}
  \city{Sun Yat-sen University}
  \country{}
  }
\email{linlng@mail.sysu.edu.cn}
\begin{abstract}
  An embodied task such as embodied question answering (EmbodiedQA), requires an agent to explore the environment and collect clues to answer a given question that related with specific objects in the scene. The solution of such task usually includes two stages, a navigator and a visual Q\&A module. In this paper, we focus on the navigation and solve the problem of existing navigation algorithms lacking experience and common sense, which essentially results in a failure finding target when robot is spawn in unknown environments.
    Inspired by the human ability to think twice before moving and conceive several feasible paths to seek a goal in unfamiliar scenes, we present a route planning method named \textbf{P}ath \textbf{E}stimation and \textbf{M}emory \textbf{R}ecalling (PEMR) framework. PEMR includes a ``looking ahead'' process, \textit{i.e.} a visual feature extractor module that estimates feasible paths for gathering 3D navigational information, which is mimicking the human sense of direction. PEMR contains another process ``looking behind'' process that is a memory recall mechanism aims at fully leveraging past experience collected by the feature extractor. Last but not the least, to encourage the navigator to learn more accurate prior expert experience, we improve the original benchmark dataset and provide a family of evaluation metrics for diagnosing both navigation and question answering modules. We show strong experimental results of PEMR on the EmbodiedQA navigation task. 
\end{abstract}


\maketitle

\section{Introduction}
\label{intro}
In recent years, there has been an increasing interest in \textbf{Embodied} \textbf{Q}uestion \textbf{A}nswering (EmbodiedQA), which involves multiple interrelated subtasks, including language understanding, scene perception, and navigation route planning. 
\begin{figure}[t]
  \centering
  \includegraphics[width=\linewidth]{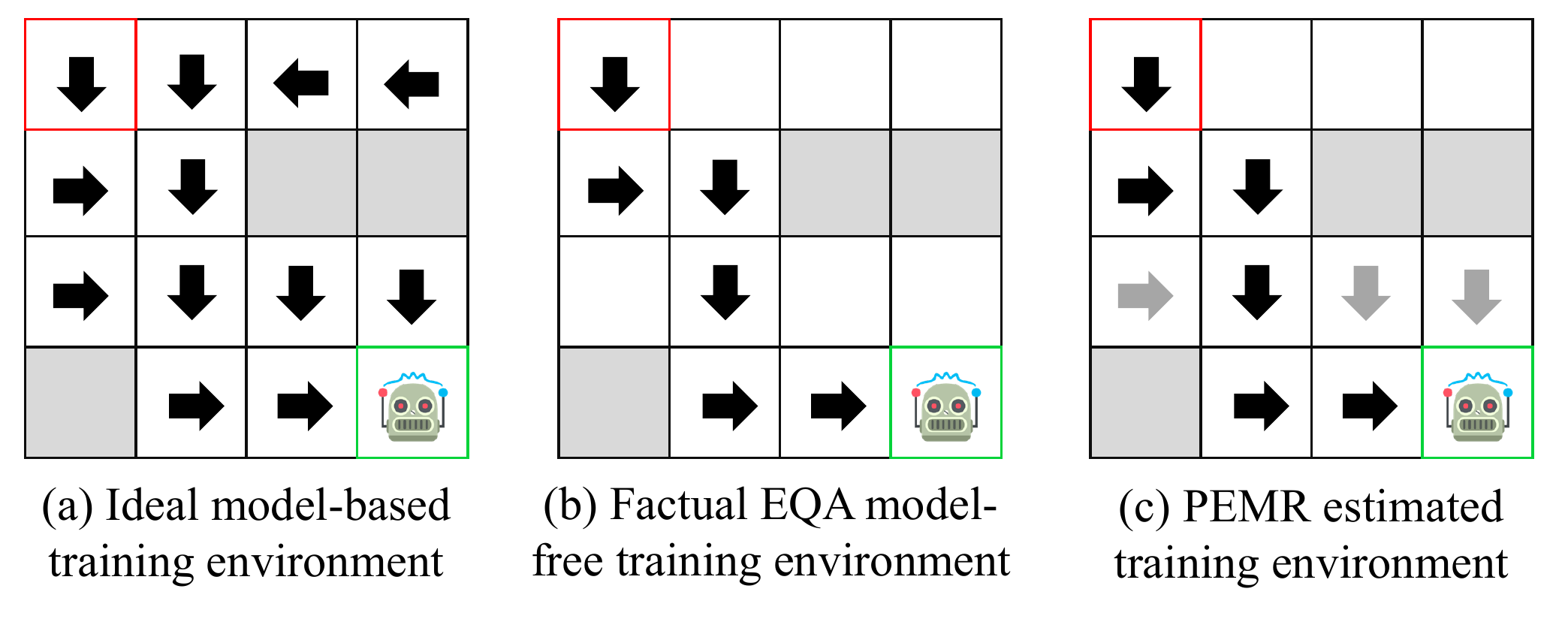}
  \caption{Assuming a 3D embodied environment is divided into accessible (white) and inaccessible (grey) grids, we simplify the EmbodiedQA procedure as: an agent robot is given a query at the outset (red framed grid); it moves grid-by-grid controlled by a navigator; it finds the target (green framed) and provides the answer. If the navigator is ideally model-based, as shown in (a), each location in the map, the robot would always take the action (black arrows) that can reach the destination in the shortest distance. However, in the factual situation displayed in (b), estimating paths other than the given shortest path is fairly hard, considering the model-free learning strategy. In the figure (c), compared with original EQA navigation model, the proposed PEMR manages to gather and memorize more environmental information (grey arrows) around the shortest path as navigation experience.\label{fig:first}}
\Description[]{}
\end{figure}
Within the EmbodiedQA task, an agent robot is required to explore the environment and answer a given question in a free-for-exploring scene~\cite{Wu2018}. Das et al.~\cite{Das2018} first simplified this task into a combination of a navigation~\cite{Zhu2017} module and a visual question answer~\cite{Antol2015} module~(overall named PACMAN). The navigation module consists of an ``eye'' and a ``brain'', that are implemented in a multitask convolutional neural network~(CNN) and a policy network, respectively. However, we empirically find that the current navigation module in PACMAN usually does not rely on the question answering process. For example, when asked ``which room is the bowl located?'', in most cases, the robot fails to enter the dining room or the kitchen, but still provides a correct answer. We argue that the above failure phenomenon mainly stems from the following two factors:

\begin{itemize}
    \item \textbf{Inadequate exploration on global visual information.}\\
     PACMAN contains a multitask CNN utilizing the feature estimator trained on semantic object segmentation and RGBD image reconstruction tasks~\cite{Wu2018}. Although the low-level visual features extracted from the CNN part are helpful for scene understanding, another component of PACMAN---the policy network---usually fails to build a close connection between the navigation goal and the first-person-view observations from visual feature estimator, since constructing this connection requires agents equipped with a global layouts view, while the provided is merely local visual information.
     Thus, we suppose the temporary model needs more exploration on the global visual information to bridge this gap.\\

    \item \textbf{Lack of exploitation on past experience.}\\
    The model PACMAN employs two possible ways to train its navigator: (1) a model-free\footnote{A model-free algorithm is an algorithm which does not use the transition probability distribution associated with the Markov decision process (MDP). The contrary concept of model-free is model-based, see Figure~\ref{fig:first}.}~\cite{sutton2018reinforcement} reinforcement learning (RL)~\cite{Mnih2016,mnih2016asynchronous}; (2) a sequential supervised learning (\emph{a.k.a} behavioral cloning (BC)~\cite{Ross2014}. However, the existing BC and RL are ineffective to recall and identify useful experience from past routing to compute gradient estimates, which leads to a bias accumulation. When this happens, the agent can barely walk on the ``correct'' path in the strange environment. The executed wrong actions will cause the environment responding with enormous negative feedbacks, which can be considered as learning classifiers upon an imbalanced dataset~\cite{kotsiantis2006handling}.
    Due to the highly complex 3D environments, this problem becomes even worse in EmbodiedQA tasks. We surmise that this problem is caused by the lack of full exploitation of the routes space when given little supervision information or positive rewards. A possible solution to overcome this problem is to discover more feasible paths with positive feedbacks. 
\end{itemize}

An ideal navigator can always find the shortest path to the target, as shown in the Figure~\ref{fig:first}(a). However, factually and especially in complex 3D environment, learning a model-free navigation function is difficult, considering the navigator is trained with limited positive data (displayed in Figure~\ref{fig:first}(b)). 
To provide more helpful visual information and tangible global experience for the navigator network to model complex environments, we attempt to add feasible paths observed by each step to the layout information storage. A {\em feasible path} is a short future path that might lead the agent to the object, represented by a symbolic action list from current observations. All feasible path features in the past can contribute to the rest policy of route planning. This mechanism is inspired by the ability of humans: when asked to find an object in strange environment, in contrast to the robots that estimate next step with highest possibility to find the target, humans usually start the searching without any global planning, but using countless feasible path selections. These selections are made based on the past experience or common sense. 

In this sense, we propose a route prediction method---\textbf{P}ath \textbf{E}stima-tion and \textbf{M}emory \textbf{R}ecalling framework (PEMR). With PEMR framework, we expand the exploring experience space by recalling ``once seen'' paths, for example, as shown in Figure~\ref{fig:first}(c), the PEMR manages to estimate more plans around the given ground-truth, which is beneficial for testing when spawn in other unseen environments. Overall, PEMR is an end-to-end learning framework that manages to utilize and recall past experience to adjust new environments. 

PEMR framework mainly consists the following three components: (a) a visual convolutional extractor that extracts the feasible path mask from the current observation; (b) a policy network that step-by-step transfers the visual features into an action list that represents the feasible path, which also serves as the experience for the next-step action prediction; and (c) an online attention mechanism that can make use of previous experience to realize the final action decision. Through combining components (a) and (b) of PEMR, we can explore more visual information in different modes that connect the navigation and partial observations; we call this a ``looking ahead'' process. The latter half process ``looking behind'' process aims at utilizing the local exploration experience in the ``looking ahead'' stage, particularly by step-by-step structuring the navigation vision searching space into a tree and transferring the space fragments into useful navigation information.
 Through iteratively acquiring access to the nodes of the tree constructed with past and current feasible paths, the learning difficulties are thereby reduced. 

Last but not the least, after investigating the existing benchmark dataset EQA v1~\cite{Das2018a}, we find some expert shortest paths are sub-optimal: the target object is missing or ill-captured from the perspective of the last few steps, which are the visual source for the question answering module. To reduce the negative influence of inaccurate prior information, those samples are either removed or modified in our experiments. To more effectively evaluate the EmbodiedQA models, we design a set of new metrics in joint consideration of both navigation and QA. We conduct experiments on both old and new datasets, and the final results show that PEMR outperforms other existing methods. 

The main contributions can be summarized as follows:
\begin{itemize}
    \item \textbf{Model Generalizability Improved.} We present an auxiliary module with path searching that can improve the versatility of the navigation model by adding global visual information provided by current and past feasible paths.\\ 
    \item \textbf{Learning Difficulties Reduced.} The model learning difficulties are reduced via a proposed action decision mechanism that fully exploits the local experience by building a structural navigation searching space and retrospectively recalling the feasible paths.\\
    \item \textbf{Dataset Reformed.} We modify the original public benchmark dataset to guarantee accurate prior information and environment feedbacks for BC and RL. A new set of evaluation metrics is introduced for EmbodiedQA diagnosed evaluation. 
\end{itemize}

\section{Related Work}
\label{sec:related_work}
\noindent\textbf{Embodied Question Answering}\\
EmbodiedQA was first introduced by Das et al.~\cite{Das2018}, who presented the PACMAN model that was initially trained with behavioral cloning and later fine-tuned by reinforcement learning~\cite{Williams1992}. PACMAN contains two components, one is the navigator that consists of a planner and a controller, and the other is a visual question answering module. Recently, Das et al. proposed another modular approach~\cite{eqa_modular}, which is a hierarchical policy network where subgoals are learned by specified subpolicies. In addition,~\cite{Yu2019} introduced a multitarget EmbodiedQA with questions that contain several target objects. Wijmans
 et al.~\cite{eqa_matterport} introduced an expansion-to-perception version EmbodiedQA by learning the navigation policies with 3D point clouds. Later Anand et al.~\cite{Anand2018} proposed a blindfold baseline of EmbodiedQA that ignores the visual information in this task. This work also highlights the gap between navigation and question answering but in an indirect way. Li et al.~\cite{Li2019} built a navigator with generative model VAE~\cite{Kingma2014} that creates future observations. Besides, a new EmbodiedQA work E2E~\cite{Wu2020} presented a landmark setting for easier and more practical implementation and achieves promising results. This work also makes contribution to improve the navigator generalizability.\\

\noindent\textbf{Visual-and-Language Navigation}\\
The recent visual-and-language navigation~\cite{Anderson2018} (VLN) is also an embodied navigation task that requires agents to follow the natural linguistic instructions in the 3D house scenes. To solve the cross-modal grounding, ill-posed feedback, and generality problems in this task, Wang et al.~\cite{Wang2019} combined the reinforced power of the learning framework with behavioral cloning. Ma et.al.~\cite{Ma2019} claimed that a regret mechanism can help make action decisions in the VLN. Considering the cross-modal property of the embodied tasks,~\cite{Ma2019a} managed to closely connect the navigator with the language instructions by utilizing a self-monitoring approach. Another language guiding navigation task~\cite{Nguyen2019} utilizes dialogue to intervene in the navigation, which makes the whole process more realistic compared to those without intervention.\\

\noindent\textbf{Reinforced Route Prediction}\\
Traditional methods for the goal-reaching visual navigation problem usually contain an execution of the route prediction. \cite{Mousavian2019} mentioned that embedding the representations of both the spatial layouts and the semantic contextual cues can improve the exploration. Study~\cite{Gupta2019} showed that spatial memory also benefits the navigation. Since RL is the elementary method to solve the route prediction problem, researchers usually evaluate newly proposed RL algorithms on route prediction tasks. For example, Altahhan~\cite{Altahhan2016} presented a self-reflection concept in deep RL and found that the offline self-reflective mechanism can boost the navigation performance. Faust et al.~\cite{Faust2018} improved the route predicting process by combining a probabilistic sampling-based mechanism with reinforcement learning. For more complex navigation environments, Lee et al.~\cite{Al-Shedivat2018} built subgoal graphs of environments for informing the RL procedure, which is akin to a hierarchical RL~\cite{Le2018} and is closely related to~\cite{eqa_modular}.

\begin{figure}[t]
    \centering
    \includegraphics[width=0.9\linewidth]{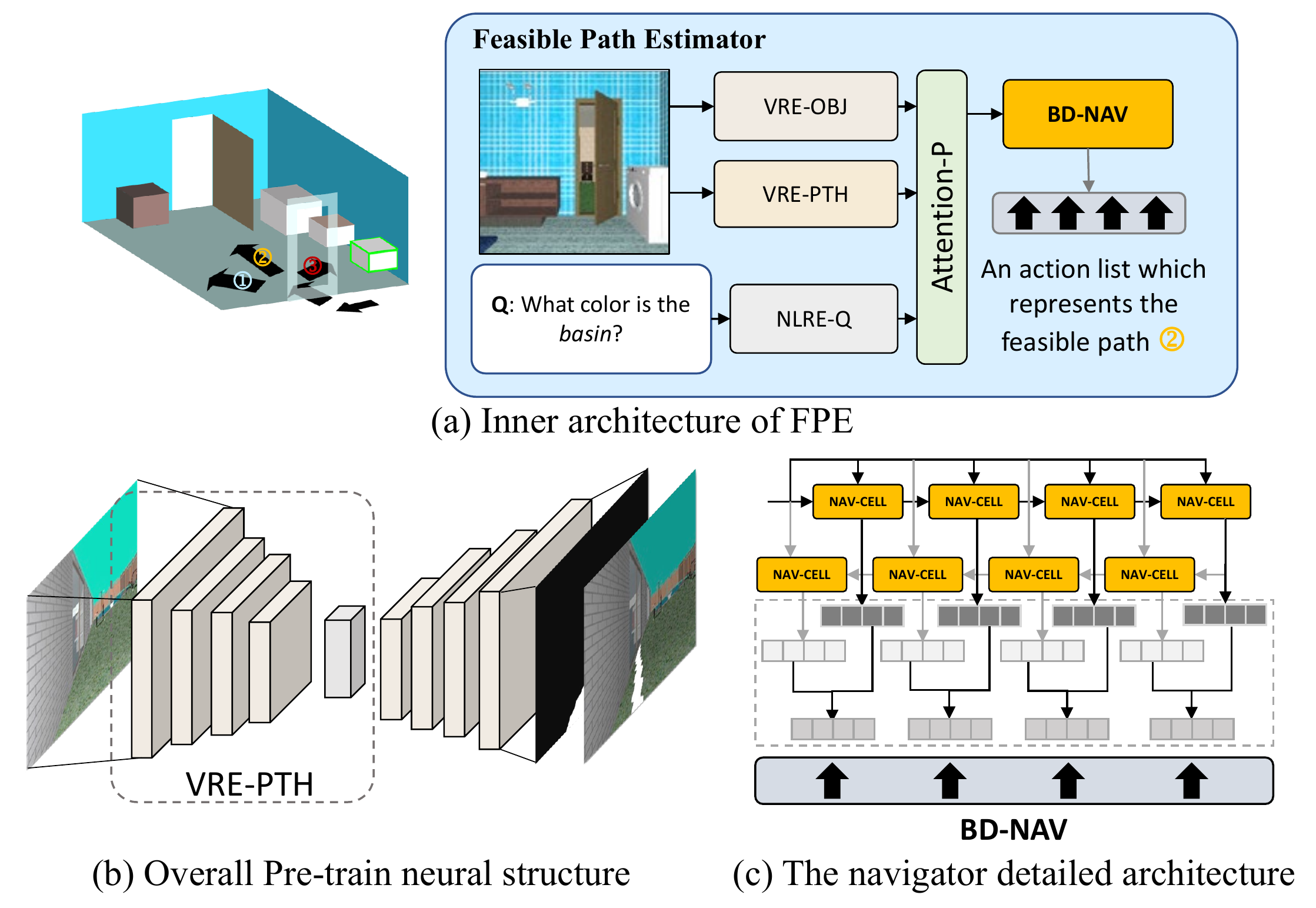}
    \caption{ This figure illustrates the implemented FPE module of ``looking ahead'' process. The feasible path searching module is formed with two components: two CNNs and a recurrent network---VRE-OBJ, VRE-PTH and NLRE-Q, as shown in (a). VRE-OBJ is pretrained with multitasks~\protect\cite{Wu2018}; VRE-PTH is pretrained with the path map construction task, as shown below in (b). The recurrent network LSTM navigator shown in (c) (in BD-NAV structure) takes a concatenated attention feature as input, which is composed of visual latent features and the embedded query vector. The output of the feasible path searching module is the symbolic action list that represents future paths.
     \label{fig:cnn_part}}
    \Description[]{}
\end{figure}

\section{Methods}
\label{sec:methods}
In this section, we first build a simple baseline navigator that can be learned with BC and RL. Next, we reformulate this baseline to construct PEMR framework. PEMR can be trained with exactly the same BC and RL specified in the baseline. Compared with the baseline navigator, PEMR invokes the visual features learned by the proposed feasible path extractor module as input (``looking ahead'' process), which is included by a recall memory machine determining actions to utilize gained experience or memory (``looking behind'' process). 
\begin{figure*}[t]
    \centering
    \includegraphics[width=0.85\linewidth]{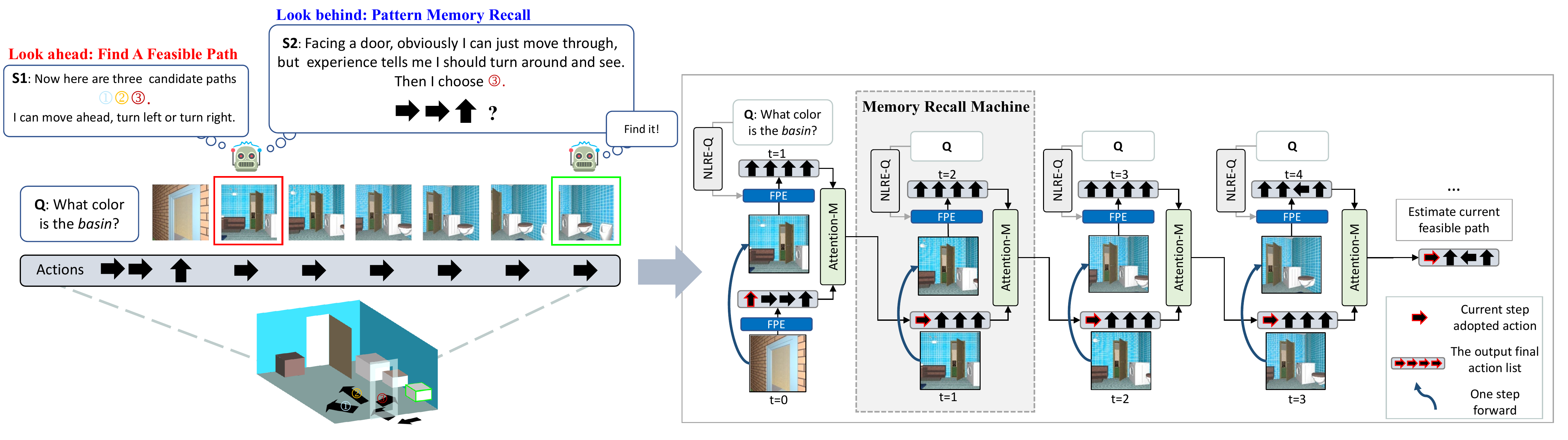}
    \caption{ The left figure illustrates PEMR navigation, including both the ``looking behind'' and ``looking behind'' processes. The right unfolds the left from time $t=0$ to $t=3$, to show how PEMR recalling and utilizing past experience. Specifically, the memory recall process follows the a path searching pattern: once the path searching module yields a symbolic action list (black arrows), a memory recall machine is activated to collect predicted actions from former steps as experience, and provides a current step decision (red framed arrows) via an attention function (Attention-M). The blue box representing FPE is illustrated in Figure~\ref{fig:cnn_part}. 
    }
    \label{fig:predictor}
    \Description[]{}
\end{figure*}
\subsection{Baseline Navigator}
\label{sec:baseline}
Suppose that in each sample, the agent carries out a sequence of actions $(a_0,a_1,...,a_T)$ and obtains a list of first-person observations $(I_0,I_1,...,I_T)$, where we use $t$ to represent each time step, and $T$ is the number of the maximum length of the navigation (the robot will be forced to stop after $T$ steps). The coordinate and horizontal degree of direction information is accessible in a specified House3D environment. There are only 4 types of movements for the robot, namely, $\forall a\in\mathcal{A}=\{\uparrow,\leftarrow,\rightarrow,\downarrow\}$, where arrows, respectively, represent moving ahead, turning left, turning right and stop. Assume that all future states of the agent are conditioned on the present and past states, and the navigation process is a stochastic process with the Markov property.

This baseline consists of a multitask CNN and a simple one-layer LSTM~\cite{Hochreiter1997} as policy network (as the navigator), which takes the convolutional features of spatially-contiguous hidden states as its input, and the navigator is learned with BC or RL. We denote question embedding as $q$, and the action embedding as $y$ ($a = \operatorname*{argmax}y$). Let $f_{\omega}$ and $\pi_{\theta}$ be the notations of the multitask CNN and the policy network, $\omega$ and $\theta$ are the parameters. Parameter $\omega$ of multitask CNN is fixed after pretraining. At each step, the agent observes $I_t$ as the input of $f_{\omega}$ and obtains the feasible path feature $f_{\omega}(I_t)$ as the output. The feed-forward process of policy network at step $t$ can be summarized as
\begin{equation}
    \label{eq:lstm_nav}
    \begin{aligned}
       &x_t=[f_{\omega}(I_t), q, y_{t-1}],\\
       &y_t,h_t =\pi_{\theta}(x_t,h_{t-1}),
    \end{aligned}
\end{equation}
where $h_t$ denotes the hidden output of the LSTM cell, and $x_t$ is the concatenated feature of visual feature $f_{\omega}(I_t)$. The question embedded feature $q$, the former action vector $y_{t-1}$ and $h_0$ are zero initialized. If $\pi_{\theta}$ is estimated through RL, then the Q-function is $Q^{\pi}(s,y)=\mathbb{E}[\sum^{T}_{t=0}\gamma^{t}R(s_t,y_t)]$, where $\gamma\in[0,1]$ is the discount factor, $R$ is the reward function, and state $s_t$ here is $[f_{\omega}(I_t),p_t]$ with the location coordinate $p$. Then, we can optimize $\pi_{\theta}$ by applying policy gradient $\nabla_{\theta}J(\theta)$, formulated as
\begin{equation}
    \label{eq:RL_J}
    \nabla_{\theta}J(\theta)=\mathbb{E}_{\pi_{\theta}}[\nabla_{\theta}\operatorname*{log}(y)Q^{\pi_{\theta}}(s,y)].
\end{equation}

Since the recurrent policy networks are usually hard to train, $\pi_\theta$ will be pretrained with BC, which is an offline supervised learning with expert suggestions, denoted by $\hat{a}_t$. A path can be represented by an action list $(a_0,...,a_T)$ with an initial position $p_0$, and we define $(\hat{a}_0,...,\hat{a}_T)$ as the shortest path to the target object in $q$. Then the parameter $\theta$ can be estimated in a supervised manner by
\begin{equation}
    \label{eq:imi}
    \operatorname*{maximize}_{\theta}\sum[\operatorname*{log}\pi_\theta(a=\hat{a}|s)].
\end{equation}
The above equation describes that $\pi_\theta$ is required to act as the ground truth under the given states. Here for simplicity, we leave out the hidden output $h$ of $\pi$, and output $y$ is the predicted action's probabilistic representation. 

\subsection{Look Ahead: Feasible Path Estimator Module}
We first introduce the visual part of PEMR --- the feasible path estimator~(FPE) module (visualized in Figure~\ref{fig:cnn_part}) --- that extracts 3D visual information from 2D observations of the environment. Although the multitask CNN (VRE-OBJ) $f_\omega$ pretrained with the depth estimation task also provides geometric information, this depth information builds a minor connection with navigation. In addition, the input visual features can be gradually neglected by $\pi$ in equation (\ref{eq:lstm_nav}), considering the policy network is a recurrent neural network~\cite{Pascanu2013}. We expect the recurrent network $\pi$ to mimic the human sense of direction to improve the navigation, based on which, we therefore add an image-to-sequence module VRE-PTH to forecast a short path at each step to the baseline as the feasible path module. This module is in advance pre-trained with shortest paths, and then will be overall learned end-to-end in the BC or RL learning stage. FPE also contains a query embedding module NLRE-Q.
We provide detailed pre-training settings in the supplementary documents~\ref{sec:feasible_pretrained}.

The formulations of PEMR will be carried out in a similar way as the baseline with subtle modifications. The FPE module is jointly represented by $f_\omega$, $\pi$ and VRE-PTH $f_{\phi}$ is another convolutional neural network parameterized with $\phi$. $f_\phi$ is the encoder of a feasible path recognition network, which is pretrained with labeled ground-truth paths, as shown in Figure~\ref{fig:cnn_part}(b), and we present the labeling method in section \ref{sec:details}. In this way, we apply VRE-OBJ $f_\omega$ to obtain multitask features and $f_{\phi}$ to obtain feasible path features. Attention-P takes concatenated $f_\phi(I_t)$, $f_\omega(I_t)$ and the embedded question representation $q$ as the input $x_t$ to the navigator. Denote $\pi^r$ as our policy network, which is an one-layer bi-directional LSTM navigator~(BD-NAV)~\cite{Zhang2015} parameterized with $\theta$. The looking forward process about $\pi^r$ can be derived by reformulating (\ref{eq:lstm_nav}) as:
 \begin{equation*}
    \label{eq:route_predictor}
    \begin{aligned}
        &x_{t}=[f_\omega(I_t), f_\phi(I_t), q],\\
        &(y_t) =\pi^r_{\theta}(y_{t-1}, x_{t}).
    \end{aligned}
\end{equation*}
The equations above show that at step $t$, given $I_t$, $\pi^r$ can generate a $k$-length\footnote{$k$ is the length of feasible path action list, here for the example $k=4$} fragment of the action probability vector $(y_t)=(y^1_t,...,y^{k}_{t})$ (the predicted path representation). The hidden output of the LSTM is substituted with the action vector $y_t$. In fact in PEMR, we substitute the LSTM of the baseline into the BD-NAV to more accurately capture the path feature but in a recurrent model with more generality. The BD-NAV module details are illustrated in the Fig.~\ref{fig:cnn_part}(c). 

\subsection{Look Behind: Patterns Memory Recall}
We depict the pattern memory recall process in Figure~\ref{fig:predictor}, where this mechanism is accomplished by a memory recall machine invoking the past and current feasible paths extracted from the FPE module. Specifically, as shown in the left part of Figure~\ref{fig:predictor}, in the ``Look behind'' box, the navigator should fully utilize past experience, namely past feasible paths provided by ``Look ahead'' process. At each time step $t$, the machine will recall $k-1$ past accessed paths to determine the next-step action, \emph{i.e.}, next step the output $a_{t+1}$ is determined by $(y_t)$, and with a unique attention function~(see the implementation details section) to fully exploit the past learned experience.

Based on the yielded feasible path in length $k$ from the FPE module, which is represented by a list of action vectors $(y_t)$, at step $t$, the memory recall machine invokes $k$ past and current feasible paths $(y_{t-k+1}),...,(y_{t})$ in the searching space. For computation efficiency, it solely evolves one predicted action from each past feasible action list and formulates a predicted action fragment $y^{k-1}_{t-k+1},...,y^1_{t-1},y_{t}$. Intuitively, the final decision $\hat y_{t}$ should be related to all predictions from previous $k$ steps, so we employ an Attention-M to achieve this goal. This attention mechanism aims at gathering as much useful information as possible for future action anticipation; thus, we have two candidate attention functions:
\begin{equation}
    \label{eq:strategies}
    \begin{aligned}
    &\textbf{A}: \hat{y_{t}}=\sum^k_{i=0}{y^i_{t+i-k+1}},\\
    &\textbf{B}: \hat{y_{t}}=\sum^k_{i=0}{w^T_{i}y^i_{t+i-k+1}},
    \end{aligned}
\end{equation}
where $w_i$ is a trainable parameter and $\hat{y_t}$ is the final decision. This process is illustrated with an example in Figure~\ref{fig:predictor}. Then we can compute $a_{t}$ as the memory  recall machine's input at step $t$:
\begin{equation}
    a_{t}=\operatorname*{argmax}_{a}\operatorname*{\textbf{softmax}}(\hat{y_{t}}), a\in\mathcal{A}.  
\end{equation}
The softmax operation within this route predictor normalizes the vector to a probability between $0$ and $1$; thus, the navigator can still be learned with RL~(\ref{eq:RL_J}) and BC~(\ref{eq:imi}). The complete framework (including feasible path estimator and memory recall attention mechanism) shares a similar network structure with the attention BD-NAV~\cite{Zhou2016}.

\section{Empirical Results and Discussion}
In the experiment section, we first provide the navigation evaluation results of PEMR compared with other existing methods. We also provide an ablation study to examine the components validity and potential of PEMR. Furthermore we provide expanded experiments in the supplementary document~\ref{sec:exp_exp}.
\subsection{Model Evaluation}
\noindent\textbf{Navigator Comparison}\\
 For a fair comparison, we provide the navigation results of $d_{\Delta}$, $d_{T}$ on both EQA v1 and EQA v1-- datasets, which are displayed in Table~\ref{tab:results_v1} and \ref{tab:results}. We obtain EQA v1-- by removing some erroneous samples from EQA v1(for details see \ref{sec:dataset}). $T_{10}$\footnote{The computation of all the metrics are specified in supplementary documents~\ref{sec:metric}} stands for backtrack 10 steps from the target as the starting position. A higher $d_{\Delta}$ implies agent can reach closer distance to the target object, and $T_{50}$ is the most difficult level. In Table~\ref{tab:results_v1}, we find PEMR has the best performance on all the $T_{30}$ and $T_{50}$ levels and obtains comparative $T_{10}$ result with MIND~\cite{Li2019} on EQA v1. But considering that at $T_{10}$ level, finding the target is quite easy: the agent starts moving quite near the target. Therefore small movements in $T_{10}$ equally result in negative and positive $d_{\Delta}$ much easier compared with $T_{30}$ and $T_{50}$.
 Hence, we consider $T_{10}$ not as quote-worthy as $T_{30}$ and $T_{50}$ to reflect the capability of route planning. 

Compared with other methods, PEMR prefers moving forward than turning circles, which makes it move to more distant place from the starting location. This can be proved by comparing results on EQA v1 and v1--. We find a decreasing in $T_{10}$ and $T_{30}$ level after removing erroneous samples from EQA v1, and this decreasing of PACMAN are much greater than that of PEMR. On the contrary, however, performances are promoted in $T_{50}$ level, and the increasing of PEMR are greater than that of PACMAN. Due to the hit accuracy of long-term navigation is the lowest, the correction on training data will seriously influence on short term and medium term navigation with high hit accuracy. Hence that phenomenon implies PEMR possesses more powerful capability on long term navigation.

Notice we have acquired inferior $T_{50}$ navigation results compared with E2E~\cite{Wu2020} on the EQA v1 this dataset (table \ref{tab:results_v1}). Referring to this work, we find the E2E framework under the original EmbodidQA task setting is simple and quite similar to our baseline (mentioned in section \ref{sec:baseline}), which is with merely a single directional LSTM as the navigator. However, since the implementation details of E2E remain unpublished and our reproduction experiment fails on $T_{50}$, if the model is built based on what the paper claims, we speculate this result can be reproduced by some random factors or techniques.\\
    \begin{table}[t]
        \small
        \centering
        \caption{Navigation results on the EQA v1 dataset. Each metric is tested under three backtrack steps. $T_{k}$ means backtrack $k$ steps starting from the target object. The default learning method is behavioral cloning (BC). The PEMR record is our best and the decision mechanism is the attention function B. We compare our results on v1 to prove its generality. Note that all the models are trained on EQA v1 and tested on its test set.\label{tab:results_v1}}
        \setlength\tabcolsep{5pt}
        \begin{threeparttable}[t]
        \begin{tabular}{ccccccc} 
            \toprule
            \multirow{2}{*}{\textbf{Method}} & \multicolumn{3}{c}{$d_\Delta\uparrow$} & \multicolumn{3}{c}{$d_{T}\downarrow$} \\
            \cmidrule(lr){2-4}  \cmidrule(lr){5-7} 
            & $T_{10}$ &$T_{30}$&$T_{50}$ & $T_{10}$ &$T_{30}$&$T_{50}$\\
            \midrule
            SegNavigation~\cite{Luo2019}&-1.22&0.15&1.62& 2.27&4.72&8.02\\
            IQA~\cite{Luo2019}&-1.69&0.15&1.70& 1.06&3.72&7.94\\
            PACMAN~\cite{eqa_modular}
            &-0.04&0.62&1.52&1.19&4.25&8.12\\
            PACMAN (BC+RL)~\cite{eqa_modular}&0.10&0.65&1.51&1.05&4.22&8.13\\
            NMC~\cite{eqa_modular}&-0.29&0.73&1.21&1.44&4.14&8.43\\
            NMC (BC+RL)~\cite{eqa_modular}&0.09&1.15&1.70& 1.06&3.72&7.94\\
            
            MIND~\cite{Li2019}&0.17&0.94&1.52&0.98&3.93&8.15\\%
            MIND (BC+RL)~\cite{Li2019}&0.25&1.21&1.65& 0.90& 3.66&8.02\\
            E2E~\cite{Wu2020}&0.23&1.48&\underline{2.36}\tnote{1}&0.92&3.39&\underline{7.39}\tnote{1}\\
            \midrule
            PEMR&\textbf{0.26}&\textbf{1.57}&1.97&\textbf{0.90}&\textbf{2.00}&7.75\\
            PEMR (BC+RL)&0.19&1.51&\textbf{2.29}&1.24&3.36&\textbf{7.46}\\

            \bottomrule
        \end{tabular}
        \begin{tablenotes}

            \item[1] \vspace{3pt} Reproduction Failure
        \end{tablenotes}
    \end{threeparttable}
    \end{table}
    \begin{table}[t]
            \caption{Results on the EQA v1-- dataset. Some mentioned in Table~\ref{tab:results_v1} are not yet public; thus, we provide only PACMAN and PEMR results on v1--. \label{tab:results}}
            \small
            \setlength\tabcolsep{4pt}
            \begin{tabular}{ccccccc} 
                \toprule
                \multirow{2}{*}{\textbf{Method}} & \multicolumn{3}{c}{$d_\Delta\uparrow$} & \multicolumn{3}{c}{$d_{T}\downarrow$} \\
                \cmidrule(lr){2-4}  \cmidrule(lr){5-7} 
                & $T_{10}$ &$T_{30}$&$T_{50}$ & $T_{10}$ &$T_{30}$&$T_{50}$\\
                \midrule
                PACMAN~\cite{eqa_modular}
                &-0.77&-0.52&1.57
                &2.41&4.30&7.43
                \\
                PACMAN (BC+RL)~\cite{eqa_modular}&-0.65&-0.46&0.95&2.29&4.20&8.04
                \\
                \midrule
                PEMR&\textbf{-0.12}&\textbf{0.81}&2.17&\textbf{1.76}&\textbf{2.97}&6.83 \\
                PEMR (BC+RL)&-0.19&0.65&\textbf{2.37}&1.75&3.15&\textbf{6.65}\\
                \bottomrule
            \end{tabular}
        \end{table}

    \begin{table*}[t]
        \centering
        \caption{This table displays $r_\Delta$, $r_T$, $o_T$, $o_\Delta$ results and question answering accuracy (Acc) tested on the EQA v1-- dataset. Our methods have better rates under all of the criteria ($T_k$), especially  $o_\Delta$, which describes the rate of missing the target. \label{tab:re_vqa}}  
        \begin{tabular}{cccccccccccccccc} 
            \toprule
            \multirow{2}{*}{\textbf{Method}} & \multicolumn{3}{c}{$r_\Delta(\%)\downarrow$} & \multicolumn{3}{c}{$r_{T}(\%)\uparrow$} &\multicolumn{3}{c}{Acc(\%)$\uparrow$} & \multicolumn{3}{c}{$o_T(\%)\uparrow$}&\multicolumn{3}{c}{$o_{\Delta}$(\%)$\downarrow$}\\
            \cmidrule(lr){2-4}  \cmidrule(lr){5-7} \cmidrule(lr){8-10} \cmidrule(lr){11-13} \cmidrule(lr){14-16}
            &$T_{10}$ &$T_{30}$&$T_{50}$ & $T_{10}$ &$T_{30}$&$T_{50}$ & $T_{10}$ &$T_{30}$&$T_{50}$& $T_{10}$&$T_{30}$&$T_{50}$& $T_{10}$&$T_{30}$&$T_{50}$\\
            \midrule
            PACMAN
            &21.9&29.7&8.4
            &75.5&47.8&14.0
            &39.9&33.8&33.6
            &22.9&12.3&4.85
            &54.1&67.1&74.9
            \\
            PACMAN (BC+RL)
            &17.4&29.6&10.8
            &79.9&48.3&13.4
            &39.9&33.8&33.3
            &21.7&14.2&5.48
            &52.1&65.2&72.8
            \\
            \midrule
            PEMR
            &16.4&\textbf{14.7}&4.0
            &80.2&62.8&\textbf{14.8}
            &47.8 & 40.5&37.0
            &24.6&14.8&5.42
            &43.3&54.7&\textbf{53.7}\\
            PEMR (BC+RL)
            &\textbf{16.1}&14.9&\textbf{2.84}
            &\textbf{84.4}&\textbf{65.5}&14.3
            &\textbf{52.2}&\textbf{45.6}&\textbf{41.4}
            &\textbf{35.9}&\textbf{20.6}&\textbf{7.67}
            &\textbf{35.1}&\textbf{46.0}&55.1\\
            \bottomrule
        \end{tabular}
    \end{table*}

\noindent\textbf{Comparison on Embodied Concepts}\\
Herein we discuss two concepts of navigation success---\textit{under-checked} and \textit{well-checked}. Whether the yielded plans are successful can be measured through $r_\Delta$, $r_T$ and $o_T$, $o_\Delta$. We provide records of these metrics tested on dataset EQA v1-- in Table~\ref{tab:re_vqa}, where we also provide the question answering accuracy. A plan is well-checked implies that agent manages to observe the target in proper location and utilize this information for correctly answering the given question; a plan is under-checked instead means with the agent once sees the object but ``forgets'' and merely provides a guessing answer in the end. High $r_T$ means high probability of entering the target room and $o_T$ measures rate of answering with the witness. If a navigator has both high $r_T$ and $o_T$, it highly possibly executes reasonable and successful navigation (well-checked). $r_\Delta$ is the rate of the agent once entering the target room but walking out before the last five steps, and $o_\Delta$ means the portion of missing target in the last five steps (no visual foundation for VQA). Thus lower $r_\Delta$ and $o_\Delta$ rates indicate that it is less likely to miss the target object (under-checked). Our PEMR has a higher $r_T$, $o_T$ while a lower $r_\Delta$, $o_\Delta$ than those of PACMAN, hence PEMR is better considering the two concepts above. Besides, $o_T$ is highly related with correct answering rate, thus PEMR also has higher VQA accuracy. 

\begin{figure}[t]
    \centering
    \resizebox{0.48\columnwidth}{!}{
    \subfigure[\Large{BC $T_{10}$}]  
    {  
        \begin{tikzpicture}
            \begin{axis}[
                width=7.5cm,
                height=5.5cm,
                xtick=\empty,
                xmin=-0.5, xmax=5.5,
                ymin=-1, ymax=2.7,
                extra x ticks={0,1,2,3,4,5},
                x tick label style={rotate=60,anchor=east},
                extra x tick labels={Baseline,PACMAN,Baseline+FPE,PACMAN+FPE, PEMR-A, PEMR-B},
                ytick={-1,-0.5,0.0,0.5,1,1.5,2,2.5},
                legend style={at={(0.0,0.4)},anchor=west},
                ymajorgrids=true,
                grid style=dashed,
            ]
            
            \addplot[
                color=blue,
                mark=square,
                ]
                coordinates {
                    (0,2.16)(1,2.41)(2,1.95)(3,1.8)(4,1.73)(5,1.69)
            
                };
                \addlegendentry{$d_T$}
            \addplot[
                color=red,
                mark=square,
                ]
                coordinates {
                    (0,-0.53)(1,-0.77)(2,-0.31)(3,-0.16)(4,-0.09)(5,-0.03)
                    
                };
                \addlegendentry{$d_\Delta$}    
            \end{axis}
            \end{tikzpicture}
    
    }  
    }
    \resizebox{0.48\columnwidth}{!}{
    \subfigure[\Large{BC+RL $T_{10}$}]  
    {  
        \begin{tikzpicture}
            \begin{axis}[
                width=7.5cm,
                height=5.5cm,
                xtick=\empty,
                xmin=-0.5, xmax=5.5,
                ymin=-1, ymax=2.7,
                extra x ticks={0,1,2,3,4,5},
                x tick label style={rotate=60,anchor=east},
                extra x tick labels={Baseline,PACMAN,Baseline+FPE,PACMAN+FPE, PEMR-A, PEMR-B},
                ytick={-1,-0.5,0.0,0.5,1,1.5,2,2.5},
                legend style={at={(0.0,0.4)},anchor=west},
                ymajorgrids=true,
                grid style=dashed,
            ]
            
            \addplot[
                color=blue,
                mark=square,
                ]
                coordinates {
                    (0,2.32)(1,2.29)(2,2.09)(3,1.9)(4,1.84)(5,1.75)
                };
                \addlegendentry{$d_T$}
            \addplot[
                color=red,
                mark=square,
                ]
                coordinates {
                    (0,-0.68)(1,-0.65)(2,-0.41)(3,-0.26)(4,-0.2)(5,-0.19)
                };
                \addlegendentry{$d_\Delta$}    
            \end{axis}
            \end{tikzpicture}
    
        }  
        }
        \resizebox{0.48\columnwidth}{!}{
        \subfigure[\Large{BC $T_{30}$}]  
        {  
            
            \begin{tikzpicture}
                \begin{axis}[
                    width=7.5cm,
                    height=5.5cm,
                    xtick=\empty,
                    xmin=-0.5, xmax=5.5,
                    ymin=-1, ymax=5,
                    extra x ticks={0,1,2,3,4,5},
                    x tick label style={rotate=60,anchor=east},
                    extra x tick labels={Baseline,PACMAN,Baseline+FPE,PACMAN+FPE, PEMR-A, PEMR-B},
                    ytick={-1,0.0,1,2,3,4,5},
                    legend style={at={(0.0,0.5)},anchor=west},
                    ymajorgrids=true,
                    grid style=dashed,
                ]
                
                \addplot[
                    color=blue,
                    mark=square,
                    ]
                    coordinates {
                        (0,3.65)(1,4.3)(2,3.24)(3,3.29)(4,3.2)(5,2.92)
                    };
                    \addlegendentry{$d_T$}
                \addplot[
                    color=red,
                    mark=square,
                    ]
                    coordinates {
                        (0,0.23)(1,-0.52)(2,0.54)(3,0.59)(4,0.58)(5,0.86)
                        
                    };
                    \addlegendentry{$d_\Delta$}    
                \end{axis}
                \end{tikzpicture}
        
        }  
        }
        \resizebox{0.48\columnwidth}{!}{
            \subfigure[\Large{BC+RL $T_{30}$}]  
            {  
                \begin{tikzpicture}
                    \begin{axis}[
                        width=7.5cm,
                        height=5.5cm,
                        xtick=\empty,
                        xmin=-0.5, xmax=5.5,
                        ymin=-1, ymax=5,
                        extra x ticks={0,1,2,3,4,5},
                        x tick label style={rotate=60,anchor=east},
                        extra x tick labels={Baseline,PACMAN,Baseline+FPE,PACMAN+FPE, PEMR-A, PEMR-B},
                        ytick={-1,0.0,1,2,3,4,5},
                        legend style={at={(0.0,0.5)},anchor=west},
                        ymajorgrids=true,
                        grid style=dashed,
                    ]
                    
                    \addplot[
                        color=blue,
                        mark=square,
                        ]
                        coordinates {
                            (0,3.86)(1,4.2)(2,3.05)(3,3.49)(4,3.16)(5,3.15)
                        };
                        \addlegendentry{$d_T$}
                    \addplot[
                        color=red,
                        mark=square,
                        ]
                        coordinates {
                            (0,-0.08)(1,-0.46)(2,0.74)(3,0.39)(4,0.63)(5,0.65)
                            
                        };
                        \addlegendentry{$d_\Delta$}    
                    \end{axis}
                    \end{tikzpicture}
            
            }  
            }
            \resizebox{0.48\columnwidth}{!}{
            \subfigure[\Large{BC $T_{50}$}]  
            {  
                \begin{tikzpicture}
                    \begin{axis}[
                        width=7.5cm,
                        height=5.5cm,
                        xtick=\empty,
                        xmin=-0.5, xmax=5.5,
                        ymin=1, ymax=8,
                        extra x ticks={0,1,2,3,4,5},
                        x tick label style={rotate=60,anchor=east},
                        extra x tick labels={Baseline,PACMAN,Baseline+FPE,PACMAN+FPE, PEMR-A, PEMR-B},
                        ytick={0,2,4,6,8},
                        legend style={at={(0.0,0.6)},anchor=west},
                        ymajorgrids=true,
                        grid style=dashed,
                    ]
                    
                    \addplot[
                        color=blue,
                        mark=square,
                        ]
                        coordinates {
                            (0,7.58)(1,7.43)(2,6.95)(3,7.06)(4,7.04)(5,6.79)
                        };
                        \addlegendentry{$d_T$}
                    \addplot[
                        color=red,
                        mark=square,
                        ]
                        coordinates {
                            (0,1.63)(1,1.57)(2,2.04)(3,1.93)(4,1.96)(5,2.21)
                        };
                        \addlegendentry{$d_\Delta$}    
                    \end{axis}
                    \end{tikzpicture}
            
            }  
            }
            \resizebox{0.48\columnwidth}{!}{
                \subfigure[\Large{BC+RL $T_{50}$}]  
                {  
                    \begin{tikzpicture}
                        \begin{axis}[
                            width=7.5cm,
                            height=5.5cm,
                            xtick=\empty,
                            xmin=-0.5, xmax=5.5,
                            ymin=0, ymax=8.5,
                            extra x ticks={0,1,2,3,4,5},
                            x tick label style={rotate=60,anchor=east},
                            extra x tick labels={Baseline,PACMAN,Baseline+FPE,PACMAN+FPE, PEMR-A, PEMR-B},
                            ytick={0,2,4,6,8},
                            legend style={at={(0.0,0.5)},anchor=west},
                            ymajorgrids=true,
                            grid style=dashed,
                        ]
                        
                        \addplot[
                            color=blue,
                            mark=square,
                            ]
                            coordinates {
                                (0,7.36)(1,8.04)(2,6.91)(3,7.21)(4,6.9)(5,6.62)
                            };
                            \addlegendentry{$d_T$}
                        \addplot[
                            color=red,
                            mark=square,
                            ]
                            coordinates {
                                (0,1.69)(1,0.95)(2,2.10)(3,1.81)(4,2.09)(5,2.34)

                            };
                            \addlegendentry{$d_\Delta$}    
                        \end{axis}
                        \end{tikzpicture}
                
                }  
                }         
    \caption{This figure illustrates the ablative testing navigation results of long-term $T_{50}$, medium-term $T_{30}$ and short-term $T_{10}$. $d_\Delta$ and $d_T$ of the baseline and PACMAN are significantly improved by adding PEMR components to the neural structure. In each subfigure, the gradually approaching red and blue lines indicate this improvement. The left charts are learned with BC and the right are learned with RL. \label{fig:ablation}}
    \Description[]{}
    \end{figure}
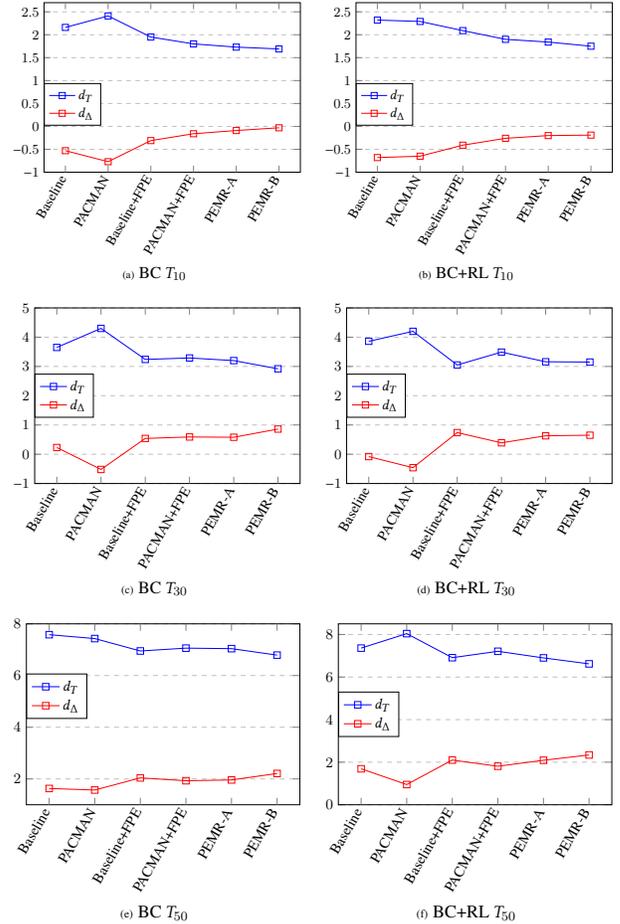 
    
\subsection{Ablation Study}
\label{sec:ablation}
We perform ablation trials to evaluate the effect of different components and learning approaches of PEMR, including the feasible path estimator(FPE), recall memory attention strategy(-A and -B) and learning strategies(BC and RL). Experimental records of navigation are illustrated in the line charts (Figure~\ref{fig:ablation}). In each sub-figure, the red line indicates $d_T$ and the blue indicates $d_\Delta$. From the charts, by gradually modifying the baseline network structure to the last framework, we observe that an obvious navigational performance growth, as in each sub-figure, the red and blue lines show a clear convergence at our model, which respectively represents less $d_T$ and greater $d_{\Delta}$.

Specifically, the PEMR can promote the model generality of navigation: all the records of PACMAN and the baseline model that have applied path module show significant improvements compared with the original model; the memory recall machine can only be implemented in a framework with the path searching module; thus, comparing the Baseline+FPE with PEMR-$*$ shows whether the recall attention machine has effects on navigation. The Baseline+FPE shows an obvious improvement on $d_{\Delta}$ compared with PEMR-A; therefore, the route predictor helps the route planning process. The records of Baseline+FPE also show that the navigator consisting of a simple LSTM has less generality, which is likely due to the simple LSTM only enrolls limited past experience recalling. Therefore after substituting the navigation model into BD-NAV (PEMR-A), the performance of $d_\Delta$ increases.

We also compared the two memory recall attention functions: defined as PEMR-A and PEMR-B. The results show attention B can handle more complex situations since it performs better on long-term and medium-term navigation, mainly due to it contains extra estimated parameters for more generality. If we compare those results from navigators separately learned with PEMR-A and PEMR-B, we notice that learning the navigator with RL has more improvements on medium-term and long term navigation than the short-term navigation. This is reasonable since the policy network in short-term navigation neglects less experience and the recall machine can make up more for the long-term and medium-term navigation. In conclusion, PEMR can effectively promote EmbodiedQA navigation, especially benefit long-term navigation.

\section{Conclusion}
In this paper, we propose a route predicting method named a path estimation and memory recalling framework (PEMR).
We build PEMR framework to improve some prevailing limitations: the weak visual model generality and lack of useful experience in RL and BC. PEMR includes two modules: 1) a feasible path searching; 2) a memory recall process where a recalling machine makes use of the past experience in the step-by-step exploration. Furthermore, in experiments, we remove some ill samples from EQA v1 and regenerate EQA v1--. We test our model on both datasets and evaluate the navigation with reasonable metrics. Consequently, experimental results show that our framework excels further navigation with a lower rate of off-targets, indicating that PEMR outperforms other existing navigators.

\bibliographystyle{ACM-Reference-Format}
\bibliography{sample-base}

\clearpage
\twocolumn[
{\Huge \bf \centering Supplementary Material \par}]
\appendix
\section{Research Methods}

\subsection{Feasible Path Module Pretrained}
\label{sec:feasible_pretrained}
We pretrain the CNN network that includes the VRE-PTH module with a feasible path mask to ensure VRE-PTH outputting visual features contains 3D geometric information of feasible path. This network is ended up with a mask-to-action fully-connected classifier, where the action is the action list that representing certain feasible path.

\begin{figure}[h]
  \centering
  \includegraphics[width=\linewidth]{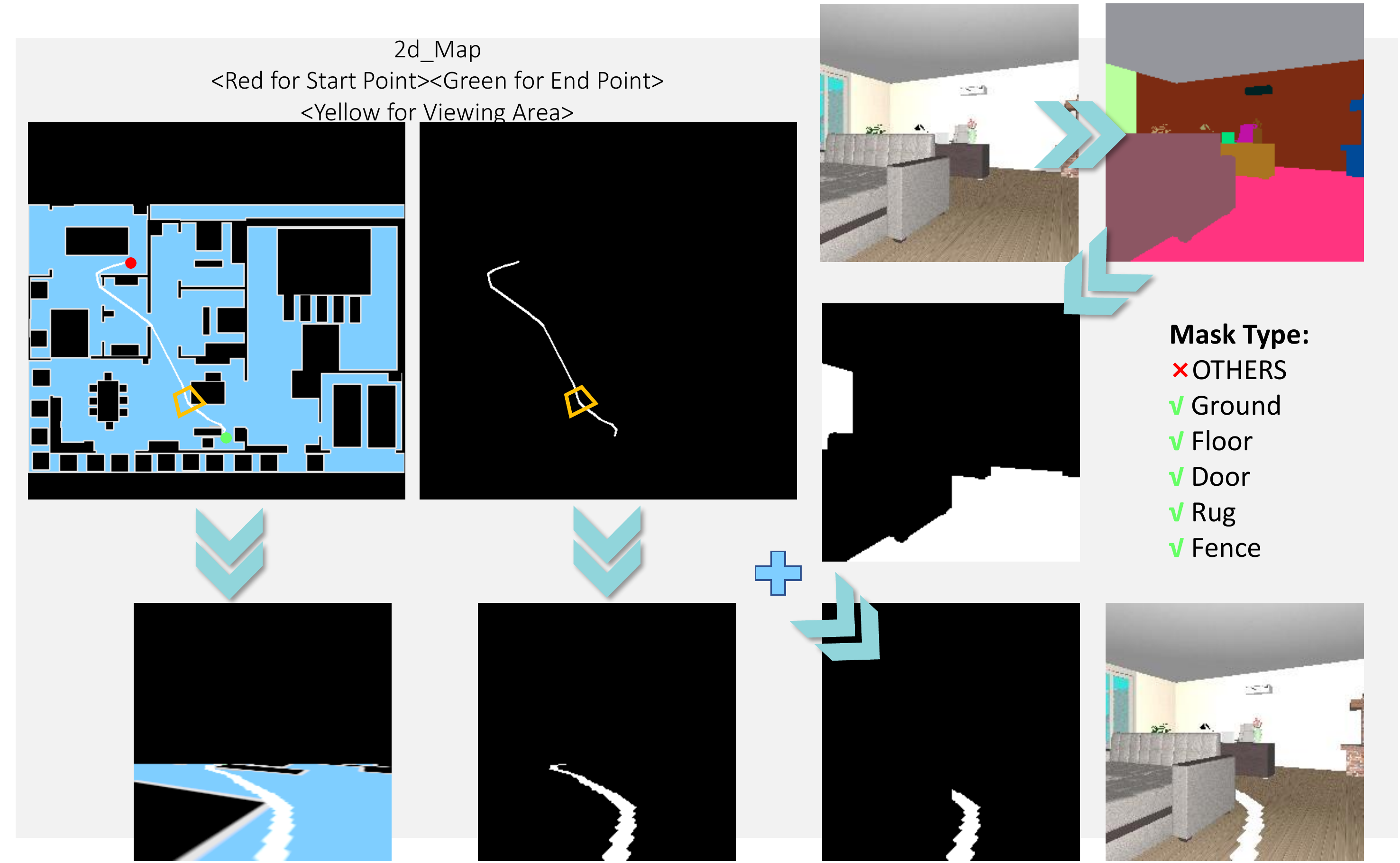}
  \caption{The generation process follows the steps from the left upper corner to the right bottom corner for the 2D global view to 3D first-person-perspective image. We place several arrows and the operators for guidance.}
  \label{fig:mask}
  \Description[]{}
\end{figure}

The feasible path mask is generated as illustrated in Fig.\ref{fig:mask} following those steps. The mask is generated through the following steps: (1) Project the shortest path from the 2D house map to 3D first-person-vision; (2) Randomly select one key location alongside the path rendering House3D and obtain the first-person-vision image with its semantic segmentation; (3) Select the segmented masks that belong to robot-physically-accessible categories, such as ``Ground'', `` Floor'' and ``Door''; (4) Do the ``and'' operation of this selected segmentation mask and the 3D projection of the path; then we obtain a ``fake 3D'' path of at this vision. We choose 100 houses and obtain nearly 100000 samples to pretrain the path searching model.

\section{Implementation Details}
\label{sec:details}
\subsection{Platform and Datasets}
\label{sec:dataset}
We implement our model on a 3D platform House3D~\cite{Wu2018} built on SUNCG~\cite{Song2017} which contains thousands of vivid 3D house environments. However, we found some shortest paths of the original dataset, \emph{i.e.}, EQA v1, may provide misleading visual information based on specified queries. 

\begin{figure}[t]
  \centering
  \subfigure[\textbf{Q}:What room is the \textbf{stereo set} located in? \textbf{A}: Bedroom.]{
  \begin{minipage}[t]{\linewidth}
  \centering
  \includegraphics[width=8cm]{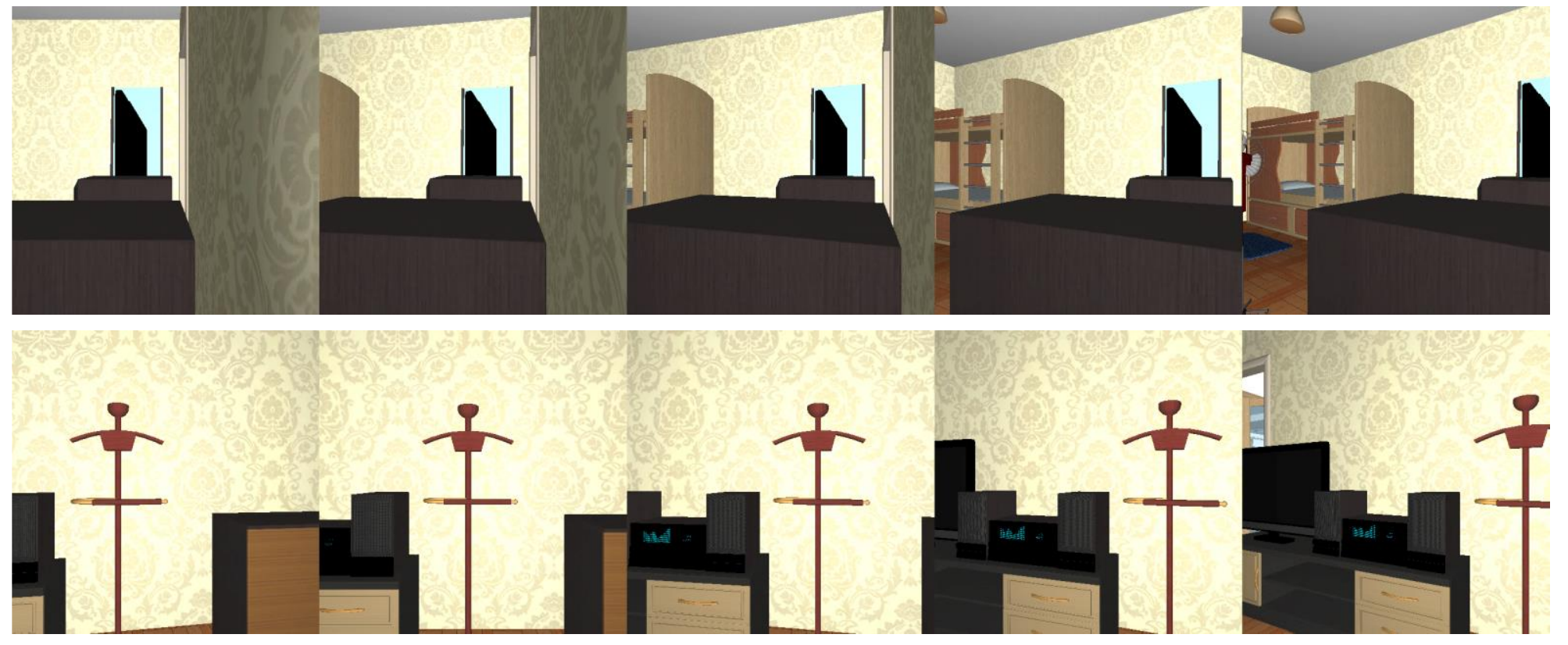}
  \end{minipage}
  }
  \subfigure[\textbf{Q}: What color is the \textbf{bunker bed}? \textbf{A}: Brown.]{
  \begin{minipage}[t]{\linewidth}
  \centering
  \includegraphics[width=8cm]{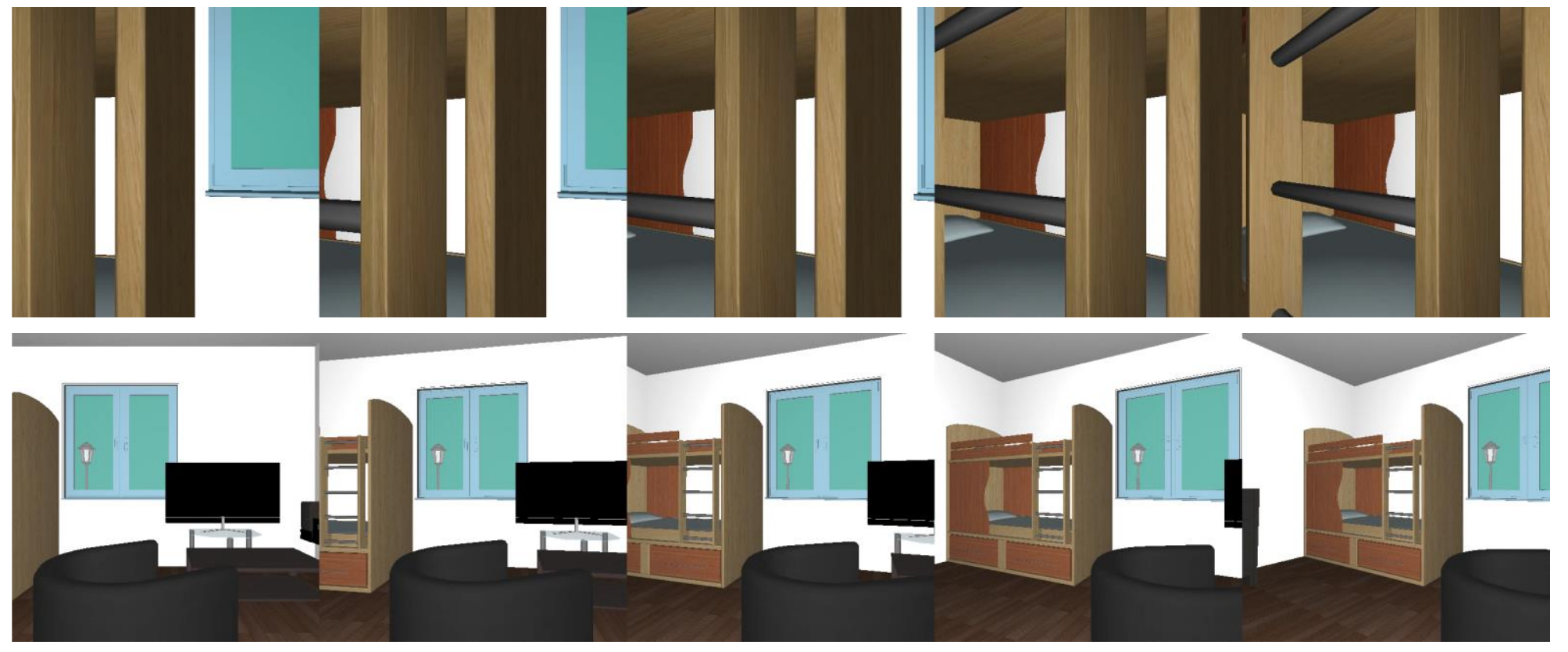}
  \end{minipage}
  }
  \caption{We visualize the last five frames of two example shortest paths under the same questions from the original dataset EQA v1 and the rectified EQA v1--. The upper row in each question is from EQA v1.  \label{fig:dataset_example}}
  \Description[]{}
  \end{figure}
  \begin{figure}[t]
      \centering
      \includegraphics[width=\linewidth]{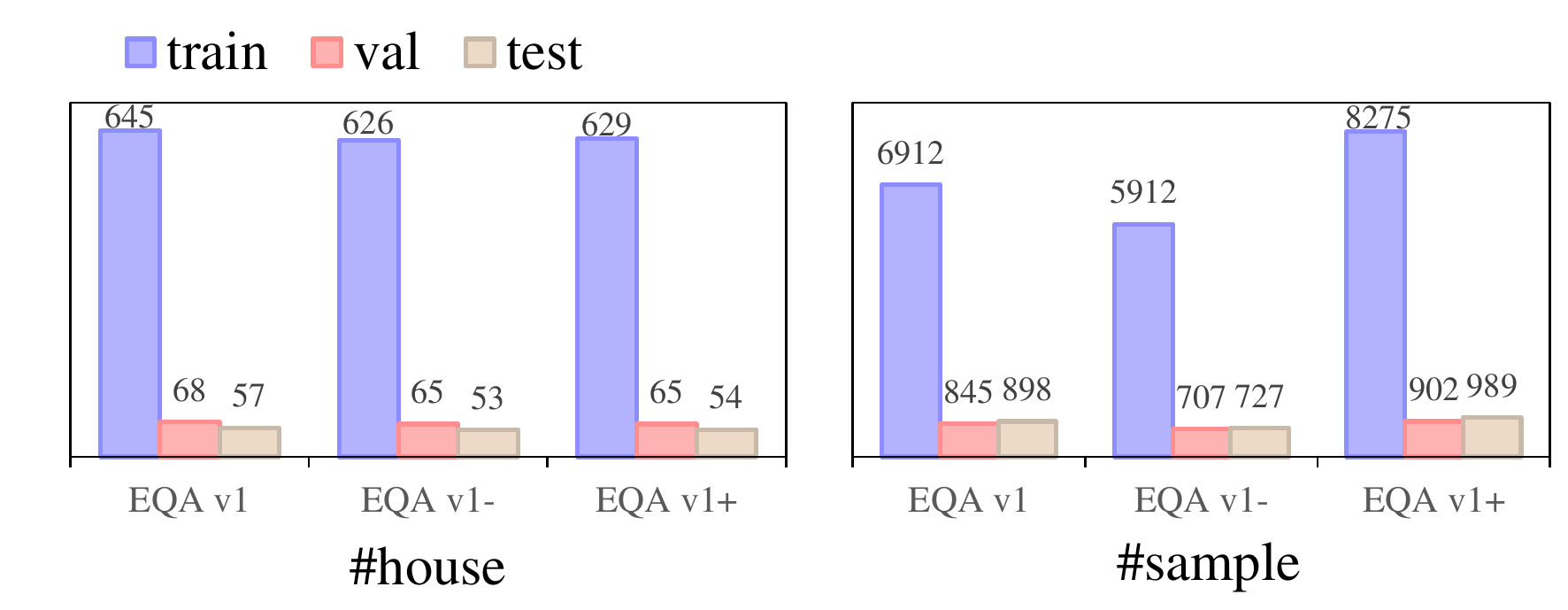}
      \caption{The statistics of both EQA v1 and our datasets. In the training set, we searched through all the samples of v1 and removed 1000 flawed ones from the training set, 200 from the testing set and obtained v1--. We further added 2000+ samples to v1-- as the new dataset v2, and the total amount of v2 is over 10 thousand. We managed to retain the data distributions, according to the shapes of the histograms.}
      \label{fig:data_stat}
      \Description[]{}
  \end{figure}
As shown in Figure~\ref{fig:dataset_example}, we compare two samples with the same query and answer from the newly generated EQA v1-- and original EQA v1 datasets. The samples from EQA v1-- have better perspectives than EQA v1, since target objects in these samples are captured in the full picture. 
Considering the last few frames will be taken as a visual source of the question answering module, we suppose that these observations should be closely correlated with the question and the answer. However, in some samples from EQA v1, the ending perspective is unreasonable, concerning the target; for instance, a stereo set is the target object of the question ``what room is the stereo set located in?'', in EQA v1, and the last five ground-truth frames only capture the top, which makes recognition difficult. In the new dataset EQA v1--, we reset all the ending positions and angles of the ill samples. Thus, in the updated dataset, the agent can observe a complete stereo set in the example sample. 

Moreover, we also deleted those samples with the incorrect shortest paths: for those from any degree of the horizontal angle, the agent can never see the target before the termination point (for example, the agent being concealed by the wall). We provide statistics of both our dataset and EQA v1 in Figure~\ref{fig:data_stat}.\\ 

\subsection{Network Structure}
PACMAN uses a multitask CNN~\cite{Wu2018} as the image feature extractor. This CNN is pretrained on the House3D platform with image segmentation, RGB to depth images and image reconstruction tasks. The complete $f_\omega$ and $f_\phi$ in the feasible path searching module are the U-Net~\cite{Ronneberger2015} structural CNNs with three convolution blocks. $f_\phi$ is trained with binary masks of the single image in the first-person-vision. The route predictor is a two-layer normal LSTM, and the attention function B of the action decision (see equation (\ref{eq:strategies})) is applied with a fully connected layer parameterized with $w$.
\\

\subsection{Loss and Reward Function}
The visual extractor $f_\phi$ of the path searching module can be equivalently regarded as a model that solves a saliency problem, where the agent manages to determine a feasible path as anticipation.  
We trained the feasible path searching module with a binary cross-entropy loss~\cite{Witten2016} function that is a sigmoid activation plus a cross-entropy loss~\cite{goodfellow2016deep}. The behavioral cloning is solved by maximizing the objective function (\ref{eq:imi}) to minimize the cross-entropy between the probability distribution of the expert data and the estimated data distribution. While we trained our model with RL, the reward function $R$ is defined as follows:
\begin{equation*}
    R(s_t,a_t) = R_1c(s_t,a_t)+R_2d(s_t,q)+R_3j(s_t,q).
\end{equation*}
$R$ is a linear combination of the collision function $c$, the distance function $d$ and the question answering precision $j$. $d$ feeds back the Euclidean distance of the agent from the current position to the target objects in the environment. The collision reward function $c$ returns a positive value if the agent successfully moves forward without any collision, a negative value if the agent collides, and $0$ when turning directions. $R_1,R_2,R_3$ are the hyperparameters, and the specific values are,  respectively, 0.5, 0.3, 0.2.\\

\subsection{Metrics and Evaluation Concepts}
\label{sec:metric}
To accurately track the agent navigation, we adopted most of the original metrics in EmbodiedQA, such as navigation metrics $d_{\Delta}$, $d_T$ and question answering accuracy, all of which have an applied special backtrack mechanism. The more steps the agent has backtracked, the greater the distance it starts from the goal place. We define short, medium and long-term navigation using this mechanism. Distance measuring is one-sided; therefore, we further define an $r_e$ as the rate that the agent used to enter the target room; an $r_T$ is defined as the rate of the last few steps when the agent is in the target room; hence, the $r_\Delta=r_e-r_T$ can be regarded as the rate of the agent failing to enter the target room.

To properly measure the learned navigator in joint consideration of the question answering module,
we introduce two evaluating concepts: the \textit{under-checked} samples and \textit{well-checked} samples, which are defined as follows: the navigation samples where the agent witnesses the given object before stopping are under-checked; those where the agent witnesses the target merely in the last five steps are the well-checked.
Evidently, the navigator with a greater output of well-checked samples is more explainable for the question answering process if the agent finally provides a correct answer.
$o_T$ denotes the rate of the target object being observed in the last five frames (the rate of well-checked success), and $o_m$ is the rate of the agent observing the object during the whole trip (the rate of under-checked success); we define an $o_\Delta = 1-\frac{o_T}{o_m}$ which describes the missed targeting rate.\\

\section{Expanded Experiments}
\label{sec:exp_exp}
In this section, we first exhibit and compare the visual feature map to explain how to estimate and determine a feasible path at each step. We next carry out an experiment to examine the model resilience ability of PEMR by testing the navigational ability on different dataset variants. Finally, we visualize some successful testing samples.
\begin{figure}[t]
    \centering
    \subfigure[Indoor Scenes]{
    \begin{minipage}[t]{\linewidth}
    \centering
    \includegraphics[width=8cm]{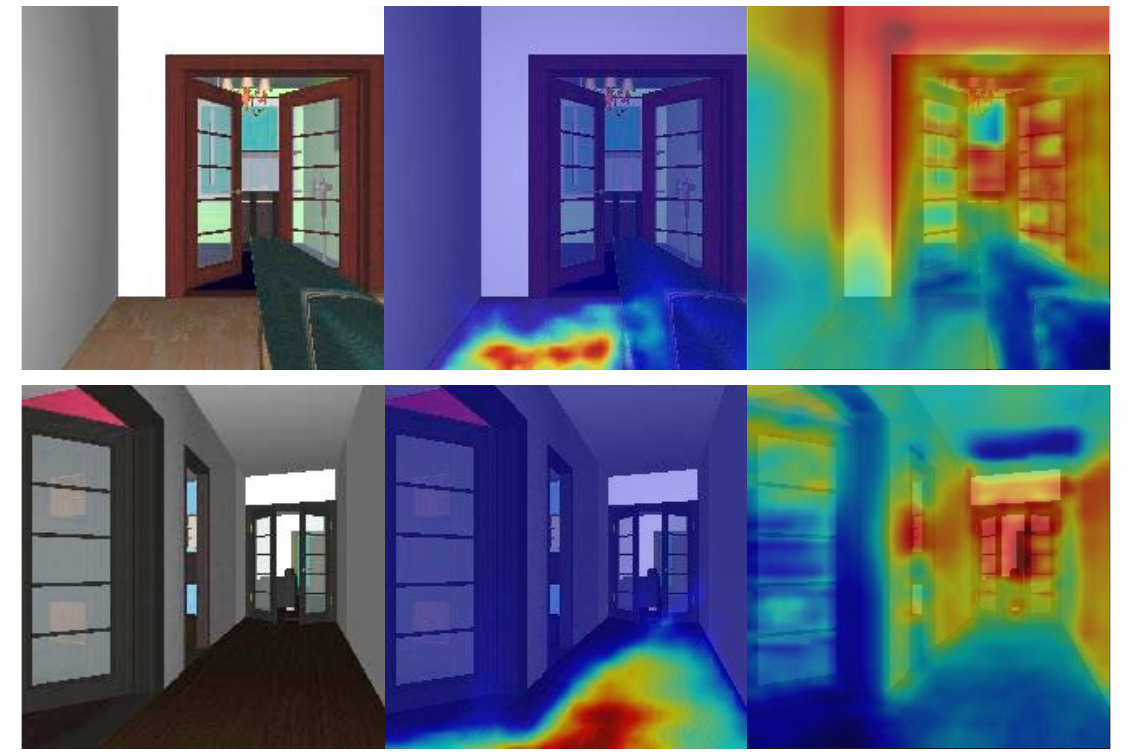}
    \end{minipage}
    }
    \subfigure[Outdoor Scenes]{
    \begin{minipage}[t]{\linewidth}
    \centering
    \includegraphics[width=8cm]{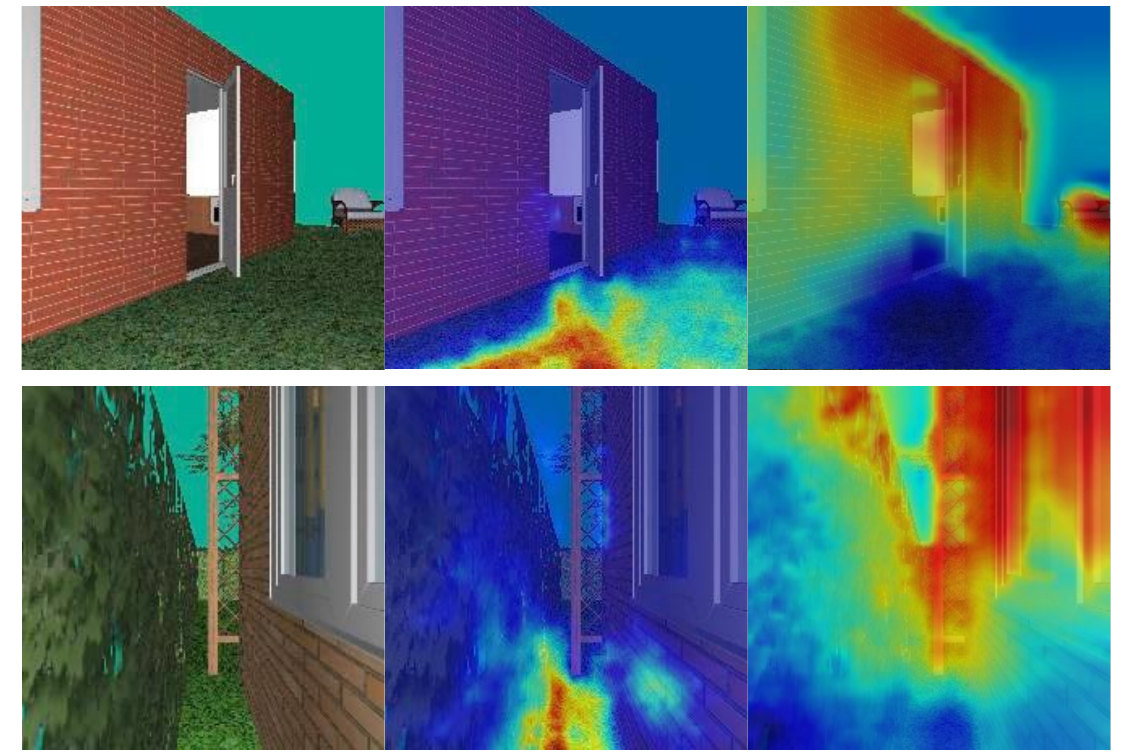}
    \end{minipage}
    }
    \caption{
    We provide a feature visualization of four samples where the heatmaps are from the feasible path searching module and the multitask CNN. As shown, regardless of whether in the indoor or the outdoor scene, the hot areas of the path mask (middle) properly capture the feasible path that is the most likely for step movement in the next time step. On the other hand, the output feature map of multitask CNN mostly captures the shapes of the objects and buildings in the scene, which might help with question answering but not with the navigation. 
    \label{fig:im_vis}}
    \Description[]{}
    \end{figure}
\subsection{Visual Information Comparison}
The path searching module is expected to learn feasible paths at each step to provide experience for agents to determine the next step movement. As shown in Figure~\ref{fig:im_vis},
based on each step observation $I$, the path searching module yields a mask $M$ (middle), where the hot area represents the most likely path. This path module is pretrained with the human-labeled mask $\hat{M}$ generated from the shortest path (the ground truth path is merely a ten-pixel length path mask, and the mask generation method is provided in the section \ref{sec:details}). We apply the mid-output feature of the mask $M$ to the route predictor and obtain an action fragment $(y_t)$ in (\ref{eq:strategies}). By acting on the actions referring to $(y_t)$, the agent can move through the predicted path. 

For comparison, we also provide the mid-output mask visualizations of the multitask CNN in Figure~\ref{fig:im_vis}, which, as shown, have high heat areas on buildings and objects. These feature maps are essentially useful since the EmbodiedQA task needs the scene comprehension information for reasonable question answering. However, these objects and building features are less helpful in navigation compared with the path masks that have a closer relationship with the route planning process. We compared the navigation model with and without this path searching module feature in the ablation study, where PACMAN with this feature shows better navigation results in Figure~\ref{fig:ablation}.\\

\subsection{Model Resilience on Dataset Variants}
\noindent In this section, we discuss the navigation ability of PEMR on different dataset variants we created for the EmbodiedQA task. We have modified EQA v1 to EQA v1-- with all ill-ended samples rectified, which will to some extent influence the navigation results. All the expansion experiment results are displayed in Table~\ref{tab:expansion}. However, we find merely some performance losses by comparing the original results with other variants; therefore, we suppose PEMR can maintain the model generality on the different navigation settings. We first test PEMR on a reversed EQA v1--. The robot will begin its trip with the camera facing the reversed direction from the original; the question and final target remain unchanged. This dataset is named EQA v1--R, and PEMR can still obtain comparative navigation results compared to v1--. We see that the effect decreases because ``turning around and walk as usual'' will take more steps and make the task more complicated. We next add up v1-- with v1--R together to see whether there is performance loss, and we obtain improvements on the v1--R compared with the v1-- results. This consequence makes sense considering PEMR has better navigation on v1--R. We finally generated 2000 more samples and combined them with the EQA v1-- to reformulate a new dataset v1$+$. The navigation results of PEMR on this v1$+$ dataset are also provided in the table.

        \begin{table}[t]
            \small
            \caption{Results of PEMR (BC+RL) evaluated on all EQA datasets variants, where EQA v1 is the original dataset; v1-- is the v1 rectified version; v1--R is the reversed v1--; v1$+$ is the v1-- dataset that consists of 2000 more newly generated samples. 
            \label{tab:expansion}}
            \begin{tabular}{ccccccc} 
                \toprule
                \multirow{2}{*}{\textbf{Dataset}} & \multicolumn{3}{c}{$d_\Delta\uparrow$} & \multicolumn{3}{c}{$d_{T}\downarrow$} \\
                \cmidrule(lr){2-4}  \cmidrule(lr){5-7} 
                & $T_{10}$ &$T_{30}$&$T_{50}$ & $T_{10}$ &$T_{30}$&$T_{50}$\\
                \midrule
                EQA v1  &0.19&1.51&2.29&1.24&3.36&7.46\\
                EQA v1--&-0.19&0.65&2.39&1.15&3.15&6.75\\
                EQA v1--R&0.08&0.89&1.89&1.56&2.89&7.10\\
                EQA v1-- \& v1--R&0.04&0.76&1.75&1.60&3.02&7.24\\
                EQA v1$+$&-0.19&0.23&1.68&1.83&3.55&7.31\\
                \bottomrule
            \end{tabular}
        \end{table}
\begin{table}[t]
    \centering
    \caption{The ablation study results are listed in this table. The values that vary in metrics after modifying the components are colored in red for the increments and green for the decrements. The following experiments are carried out on dataset EQA v1-. In this table we evaluate the effects of our components---the feasible path extractor~(FPS) module and action decision strategy. The path module can be implemented on any navigator as a visual feature extractor in the whole network framework, while the action decision can only be used in our particular navigator; thus, we merely compare the results of two strategies (plan A and B) here.\label{tab:ablation}}
    \setlength\tabcolsep{1pt}
    \small
    \begin{tabular}{lcccccc} 
    
      \toprule
      \multirow{2}{*}{\textbf{Model \& Module}} & \multicolumn{3}{c}{$d_\Delta\uparrow$} & \multicolumn{3}{c}{$d_{T}\downarrow$} \\
      \cmidrule(lr){2-4}  \cmidrule(lr){5-7} 
      & $T_{10}$ &$T_{30}$&$T_{50}$ & $T_{10}$ &$T_{30}$&$T_{50}$\\
      \midrule
      &\multicolumn{6}{c}{Behavioral Cloning}\\
      \midrule
      Baseline&-0.53&0.23&1.64&2.16&3.65&7.36\\
      Baseline+\textbf{FPE}&-0.31\scriptsize{\textcolor{red}{+0.22}}&0.54\scriptsize{\textcolor{red}{+0.31}}&2.08\scriptsize{\textcolor{red}{+0.44}}&1.95\scriptsize{\textcolor{green}{-0.21}}&3.24\scriptsize{\textcolor{green}{-0.31}}&6.91\scriptsize{\textcolor{green}{-0.44}}\\
      \midrule
      PACMAN&-0.77&-0.52&0.95&2.41&4.3&8.04\\
      PACMAN+\textbf{FPE}&-0.16\scriptsize{\textcolor{red}{+0.61}}&0.58\scriptsize{\textcolor{red}{+1.10}}&1.81\scriptsize{\textcolor{red}{+0.76}}&1.8\scriptsize{\textcolor{green}{-0.61}}&3.2\scriptsize{\textcolor{green}{-1.10}}&7.28\scriptsize{\textcolor{green}{-0.76}}\\
      \midrule
      PEMR-\textbf{A}&-0.09&0.58&2.09&1.73&3.2&6.9\\
      PEMR-\textbf{B}&\textbf{-0.03}\scriptsize{\textcolor{red}{+0.06}}&\textbf{0.86}\scriptsize{\textcolor{red}{+0.24}}&\textbf{2.18}\scriptsize{\textcolor{red}{+0.09}}&\textbf{1.69}\scriptsize{\textcolor{green}{-0.04}}&\textbf{2.92}\scriptsize{\textcolor{green}{-0.28}}&\textbf{6.82}\scriptsize{\textcolor{green}{-0.08}}\\
      \midrule
      &\multicolumn{6}{c}{Behavioral Cloning + Reinforcement Learning}\\
      \midrule
      Baseline&-0.68&-0.08&1.69&2.32&3.86&7.36\\
      Baseline+\textbf{FPE}&-0.41\scriptsize{\textcolor{red}{+0.27}}&\underline{\textbf{0.74}}\scriptsize{\textcolor{red}{+0.82}}&2.10\scriptsize{\textcolor{red}{+0.44}}&2.09\scriptsize{\textcolor{green}{-0.23}}&\underline{\textbf{3.05}}\scriptsize{\textcolor{green}{-0.81}}&6.91\scriptsize{\textcolor{green}{-0.45}}\\
      \midrule
      PACMAN&-0.65&-0.46&0.95&2.29&4.2&8.04\\
      PACMAN+\textbf{FPE}&-0.26\scriptsize{\textcolor{red}{+0.39}}&0.39\scriptsize{\textcolor{red}{+0.85}}&1.75\scriptsize{\textcolor{red}{+0.8}}&1.81\scriptsize{\textcolor{green}{-0.38}}&3.49\scriptsize{\textcolor{green}{-0.71}}&7.21\scriptsize{\textcolor{green}{-0.83}}\\
      \midrule
      PEMR-\textbf{A}&-0.2&0.63&2.19&1.84&3.16&6.9\\
      PEMR-\textbf{B}&\textbf{-0.19}\scriptsize{\textcolor{red}{+0.01}}&0.65\scriptsize{\textcolor{red}{+0.0.02}}&\textbf{2.34}\scriptsize{\textcolor{red}{+0.15}}&\textbf{1.75}\scriptsize{\textcolor{green}{-0.01}}&3.15\scriptsize{\textcolor{green}{-0.0.01}}&\textbf{6.75}\scriptsize{\textcolor{green}{-0.15}}\\
      \bottomrule
  \end{tabular}
\end{table}

\begin{figure}[t]
  \centering
  \subfigure[\textbf{Q}: What room is the vacuum cleaner located in? \textbf{A}: Garage.]{
  \begin{minipage}[t]{\linewidth}
  \centering
  \includegraphics[width=8cm]{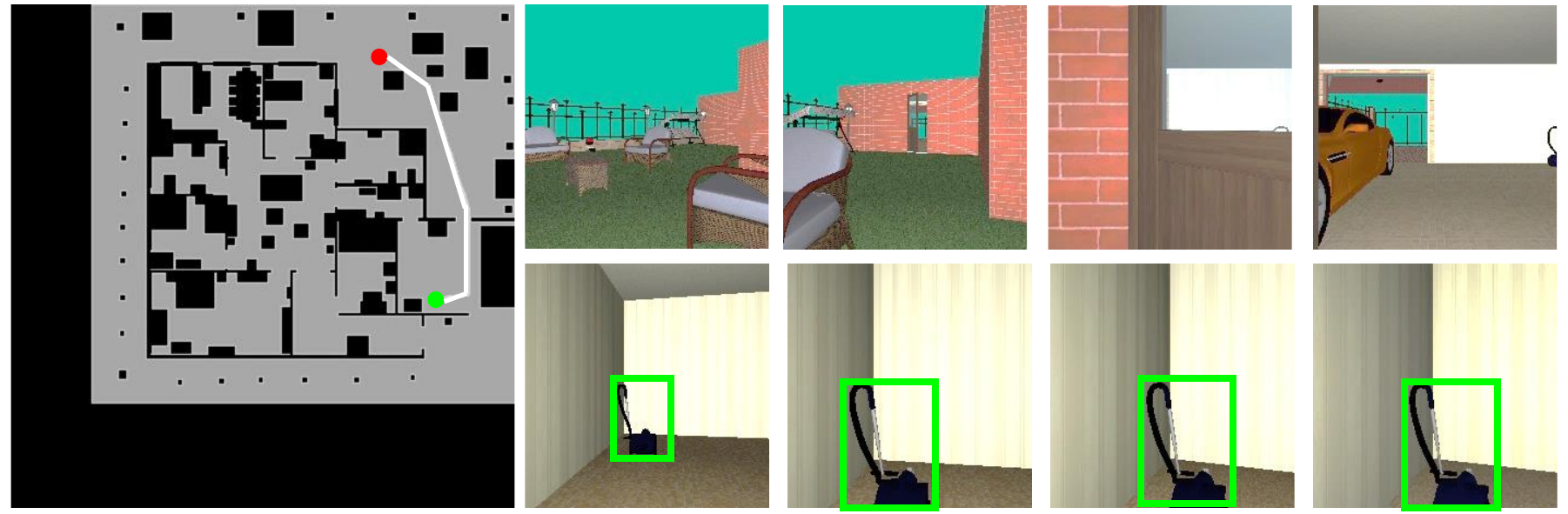}
  \end{minipage}
  }
  \subfigure[\textbf{Q}: What room is the sofa locate in? \textbf{A}: Living room.]{
  \begin{minipage}[t]{\linewidth}
  \centering
  \includegraphics[width=8cm]{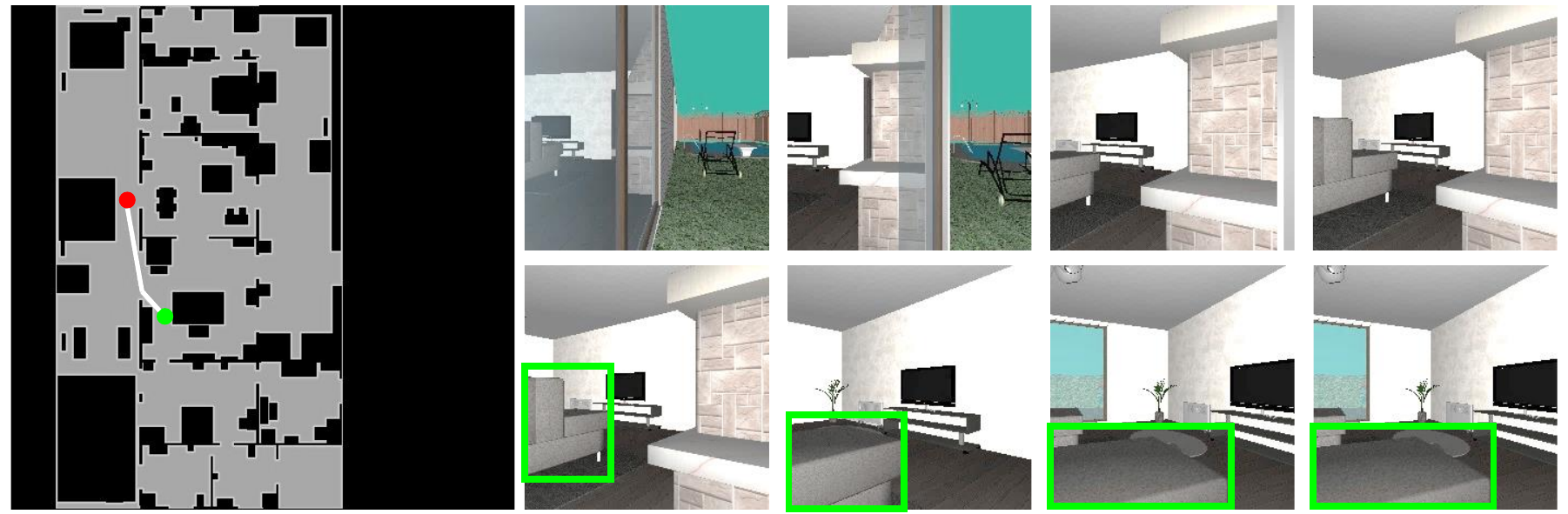}
  \end{minipage}
  }
    \subfigure[\textbf{Q}: What color is the piano? \textbf{A}: Black.]{
  \begin{minipage}[t]{\linewidth}
  \centering
  \includegraphics[width=8cm]{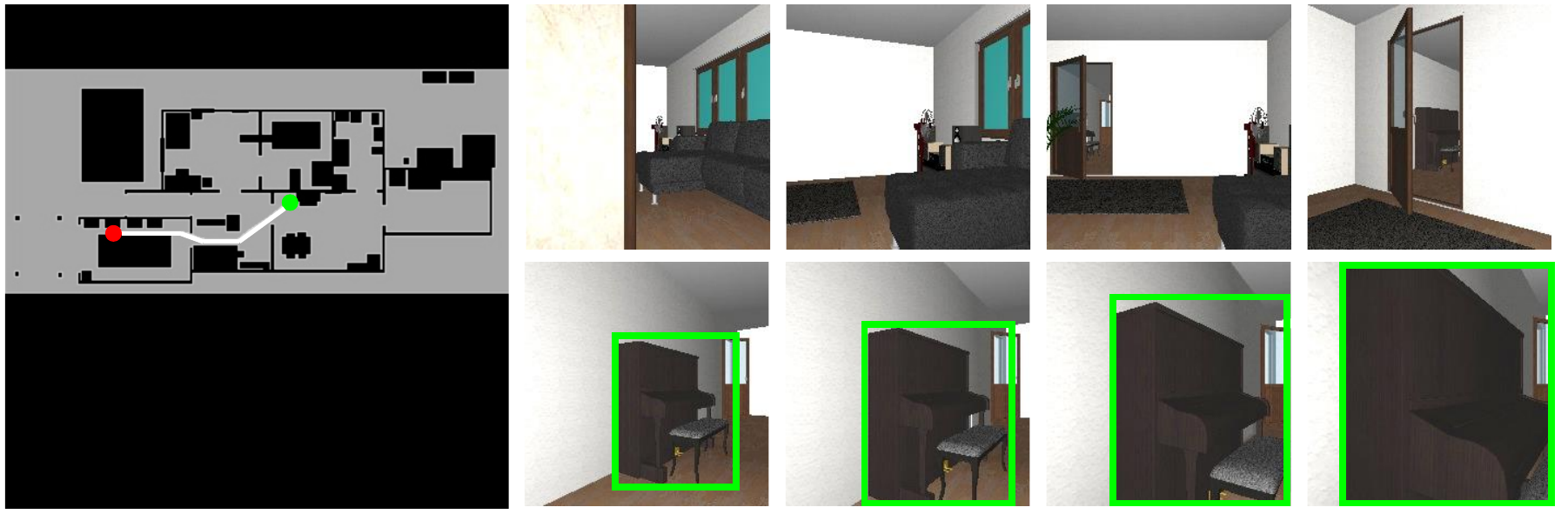}
      \end{minipage}
    }
       \subfigure[\textbf{Q}: What color is the sink? \textbf{A}: Grey.]{
  \begin{minipage}[t]{\linewidth}
  \centering
  \includegraphics[width=8cm]{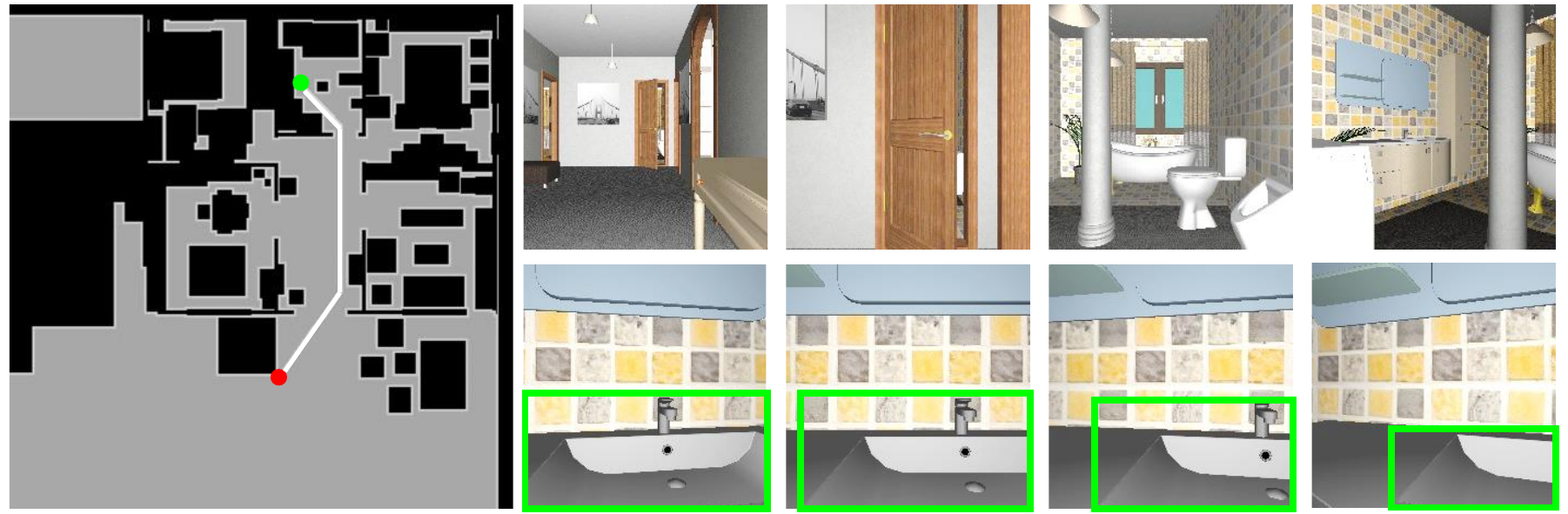}
      \end{minipage}
      }
  \caption{Visualizations of four successful testing samples. Each line of images contains a global map of the house in the left, where a white line representing the complete route, red and green spot respectively represent the start and end; first row images are the keyframes in first-person-perspective of the whole trip; the second row displays the last four frames of the navigation video, and we frame the target object with a green bounding box in this row. Capital letters \textbf{Q} and \textbf{A} stand for the query and the answer.\label{fig:vis}}
  \Description[]{}
  \end{figure}
\subsection{Effects of Proposed Components}
We unfold the experimental records of Fig.~\ref{fig:ablation} into Table~\ref{tab:ablation}, to more accurately ensure whether the components are functioning and positively influence the performance. First, the path searching module, which can also be regarded as a visual feature extractor, is significantly improving the model, even under different navigation modules. As we have mentioned in the model construction section, to implement the intuitive features, we directly concatenated the output features with the earlier visual features of the multitask CNN. We use exactly the same method to combine other EmbodiedQA navigators, and find the improvements are retained on both our baseline and PACMAN navigation frameworks. 

As shown in the Table~\ref{tab:ablation}, we can see in three difficulties (after testing under different backtrack steps $T_{10}, T_{30}, T_{50}$), the feasible path module will bring pretty remarkable improvements. For example, as we have colored the increments of $d_\Delta$ in red, the records of all the methods show red in $d_\Delta$ and green in $d_T$ after we apply them with feasible path model, under both the circumstances of only being trained with BC and being trained with BC+RL. Moreover, we suppose that with an action decision strategy, the performance can be further promoted. This conclusion can be easily drawn in the behavioral cloning learning condition, but in the bottom part of Table~\ref{tab:ablation}, where the all the network frameworks are trained with reinforcement learning after BC, we notice that Baseline+FPE has the best records of $T_{30}$ (the records are underlined). 
We suppose this phenomenon results from the random property of LSTM, which cannot simultaneously fit middle-term navigation and long-term navigation. As shown in all the tables, we find that it is hard to train a navigator to work well in all the levels of the backtrack distance. For another example, PACMAN+FPE surpasses Baseline+FPE in $T_{10}$ and $T_{30}$ but is surpassed in $T_{50}$. 

On the other hand, we notice that PEMR with BC+RL merely has better $T_{50}$ navigation than PEMR with BC. This is because training the navigator with BC will make the robot imitate the best path we select as the ground truth, while RL will make the robot act for more positive environmental feedback. Therefore, the RL-learned navigator is more active such that it can come closer to the target and has better performance on $T_{50}$, since in the long-distance navigation, and the environment will easily provide instance and consecutive positive feedback. While in the short-term ($T_{10}$) and medium-term ($T_{30}$) navigation, the ground-truth paths of testing are highly alike to those of training, which results in BC-learned navigators that have better results on these terms of navigation than RL.

\subsection{Predicted Route Visualization}
We pick some well-checked successful observation samples (finally observing the target object) for visualization. Each row in Figure~\ref{fig:vis} contains the route map, four keyframes selected from the whole trip, and the last four frames of the navigation video. The samples are those that agents start the trip at backtrack $50$ steps from its ground-truth termination. For a clear perspective, we have turned the images $0.4$ times lighter. 

\end{document}